\documentclass{egpubl}
\usepackage{eg2018}

\ConferenceSubmission   
 \electronicVersion 

\ifpdf \usepackage[pdftex]{graphicx} \pdfcompresslevel=9
\else \usepackage[dvips]{graphicx} \fi

\PrintedOrElectronic

\usepackage{t1enc,dfadobe}

\usepackage{egweblnk}
\usepackage{cite}

\title[Watch to Edit]%
      {Watch to Edit: Video Retargeting using Gaze} 
 \author[K. Rachavarapu, M. Kumar, V. Gandhi \& R. Subramanian]
{\parbox{\textwidth}{\centering Kranthi Kumar Rachavarapu$^{1}$\thanks{Both authors contributed equally to the work} , Moneish Kumar$^{1}$$^{\dagger}$, Vineet Gandhi$^{1}$
        and Ramanathan Subramanian$^{2}$ }
        \\
{\parbox{\textwidth}{\centering $^1$ CVIT, IIIT Hyderabad, $^2$ University of Glasgow, Singapore
       }
}
}

 \usepackage{mathtools}
\usepackage{eurosym}
\usepackage{mathptmx}
\usepackage{txfonts}
\usepackage{textcomp}
\usepackage{amstext}
\usepackage{float}

\def\eg{\textit{e.g.}}
\def\ie{\textit{i.e.}}

\def\etal{\textit{et al.}}
\def\etc{\textit{etc.}}

\begin{document}


\maketitle

\begin{figure*}[h!]
  \centering
  \includegraphics[width=\textwidth]{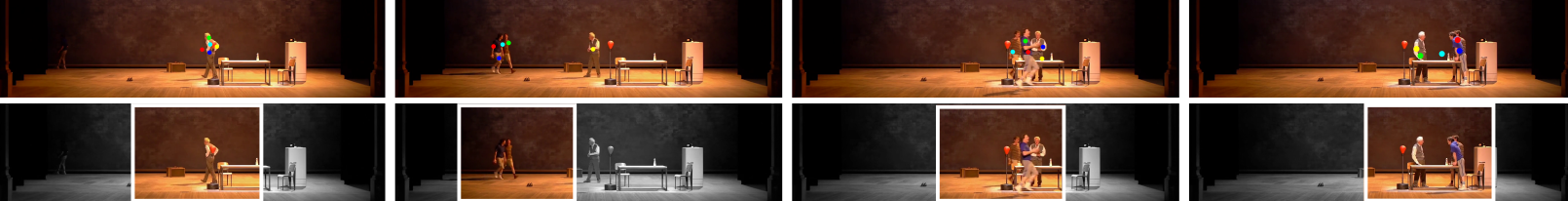}
  \vspace*{-1.9em}
  \par \hfill {\fontsize{5}{4}\selectfont {\copyright Celestin, Theatre de Lyon }}\\ 
  \vspace*{0.2em}
  \includegraphics[width=\textwidth]{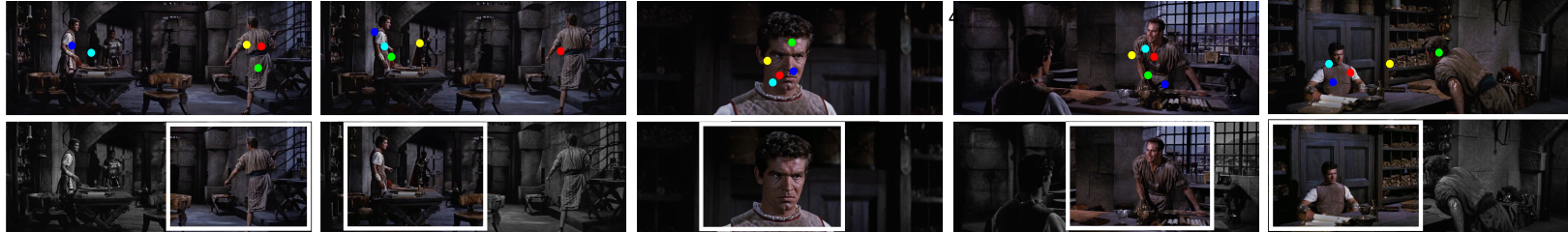}
  \vspace*{-1.8em} \par \hfill {\fontsize{5}{4}\selectfont {\copyright Warner Bros.}} \\
  
  \caption{We present an algorithm to retarget a widescreen recording to smaller aspect ratios. The original recording with overlaid eye gaze data from multiple users (each viewer is a unique color) and the results computed by our algorithm (white cropping window) are shown. Our algorithm is content agnostic and can be used to edit a theatre recording from a static camera (top) or re-edit a movie scene (bottom). }~\label{fig:intro}
\end{figure*}

\begin{abstract}
   We present a novel approach to {optimally retarget videos for varied displays} with differing aspect ratios by preserving salient scene content discovered via eye tracking. Our algorithm performs editing with cut, pan and zoom operations by optimizing the path of a \textit{cropping window} within the original video while seeking to (i) preserve salient regions, and (ii) adhere to the principles of cinematography. Our approach is (a) \textit{content agnostic} as the same methodology is employed to re-edit a wide-angle video recording or a close-up movie sequence captured with a static or moving camera, and (b) \textit{independent of video length} and can in principle re-edit an entire movie in one shot.

Our algorithm consists of two steps. The first step employs gaze transition cues to detect time stamps where new \textit{{cuts}} are to be introduced in the original video via dynamic programming. A subsequent step optimizes the cropping window path (to create \textit{{pan}} and \textit{{zoom}} effects), while accounting for the original and new cuts. The cropping window path is designed to include maximum gaze information, and is composed of piecewise constant, linear and parabolic segments. It is obtained via $L(1)$ regularized convex optimization which ensures a smooth viewing experience. We test our approach on a wide variety of videos and demonstrate significant improvement over the state-of-the-art, both in terms of computational complexity and qualitative aspects. A study performed with 16 users confirms that our approach results in a superior viewing experience as compared to  gaze driven re-editing~\cite{jain2014} and letterboxing methods, especially for wide-angle static camera recordings.
 \\

\ccsdesc[300]{Computing methodologies~Scene Understanding}
\ccsdesc[300]{Computing methodologies~Image-based rendering}
\ccsdesc[300]{Theory of computation~Dynamic programming}
\ccsdesc[300]{Theory of computation~Convex optimization}
 \printccsdesc   

\end{abstract}  
\section{Introduction}

The phenomenal increase in multimedia consumption has led to the ubiquitous display devices of today such as LED TVs, smartphones, PDAs and in-flight entertainment screens. While viewing experience on these varied display devices is strongly correlated with the display size, resolution and aspect ratio, digital content is usually created with a target display in mind, and needs to be manually re-edited (using techniques like pan-and-scan) for effective rendering on other devices. Therefore, automated algorithms which can \textit{retarget} the original content to effectively render on novel displays are of critical importance.

Retargeting algorithms can also enable content creation for non-expert and resource-limited users. For instance, small/mid level theatre houses typically perform recordings with a wide-angle camera covering the entire stage as costs incurred for professional video recordings are prohibitive (requiring a multi-camera crew, editors \etc). Smart retargeting and compositing ~\cite{wiced2017} can convert static camera recordings with low-resolution faces into professional-looking videos with editing operations such as \textit{pan}, \textit{cut} and \textit{zoom}. 

Commonly employed video retargeting methods are non-uniform scaling (squeezing), cropping and letterboxing~\cite{Shamir:2009}. However, squeezing can lead to annoying distortions; letterboxing results in large portions of the display being unused, while cropping can lead to the loss of scene details. {Several efforts have been made to automate the retargeting process, the early work by Liu and Gleicher~\cite{LG06} posed retargeting as an optimization problem to select a cropping window inside the original recording. Other advanced methods like content-aware warping, seam carving then followed~\cite{Vaquero2010}. However, most of these methods rely on \textit{bottom-up} saliency derived from computational methods which do not consider high-level scene semantics such as emotions, which humans are sensitive to~\cite{katti2010making,Subramanian2014}.} Recently, Jain \etal~\cite{jain2014} proposed Gaze Driven Re-editing (GDR), which preserves human preferences in scene content without distortion via user gaze cues and re-edits the original video introducing novel cut, pan and zoom operations. However, their method has limited applicability due to a) extreme computational complexity and b) the hard assumptions made by the authors regarding the video content-- \eg, the authors assume that making more than one cut per shot is superfluous for professionally edited videos, but this assumption breaks down when the original video contains an elongated (or single) shot as with the aforementioned wide-angle theatre recordings. Similarly, their re-editing cannot work well when a video contains transient cuts or fast motion.      

To address these problems, we propose a novel retargeting methodology via gaze based re-editing employing convex optimization. {Our work is inspired from GDR~\cite{jain2014} and also estimates a cropping window path traversing through the original video introducing pan, cut and zoom operations in the process (Figure~\ref{fig:intro})}. However, our advantages with respect to GDR are that: (1) Our convex optimization framework guarantees a feasible solution with minimal computational complexity; our method requires 40 seconds to edit a 6-minute video, whereas GDR takes around 40 minutes to edit a 30 second video due to computation of non-uniform B-spline blending; (2) Our optimization is $L(1)$ regularized, \ie, it ensures sparse editing motions mimicking professional camera capturing and optimizing viewing experience, whereas spline blending may result in small, unmotivated movements; (3) {Our method is agnostic to both video content and video length and it extends GDR to long uncut videos like theatre or surveillance recordings captured with static, large field-of-view cameras.}    

Once potentially important scene content is captured via gaze data compiled from a few users, our algorithm performs re-editing of the gaze-tracked video in two steps. The first step employs user gaze transitions to identify time stamps in the original video where novel cuts can be introduced via dynamic programming-based optimization. It also estimates a cropping window path within the original recording, such that the cropping window encompasses maximum gaze information. The next step employs an $L(1)$ regularized convex optimization to convert this path into a professional looking one. The $L(1)$ regularization models the virtual camera trajectory via piecewise static, linear and parabolic segments, and ensures that the virtual camera moves frugally and fluidly akin to a professional camera, thereby resulting in a smooth viewing experience. This is confirmed via a user study involving 16 users, where our approach is perceived as enabling optimized viewing of scene emotions and actions, as compared to gaze based re-editing~\cite{jain2014} and letterboxing especially for static theatre recordings. 

The next section overviews related work, in order to highlight the salient aspects of our approach. The following sections provide a detailed discussion of our retargeting approach, associated experiments and the user study. 

\begin{figure*}[h!]
  \centering
  \begin{tabular}[b]{c c c}
  \includegraphics[width=.3\textwidth]{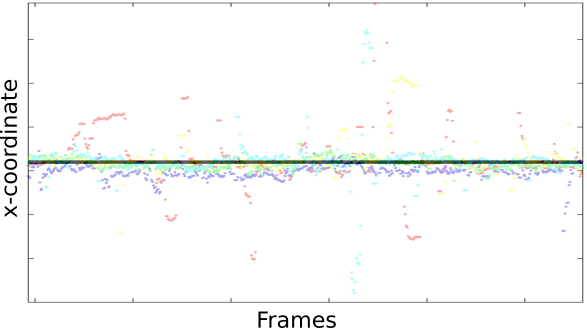} &
  \includegraphics[width=.3\textwidth]{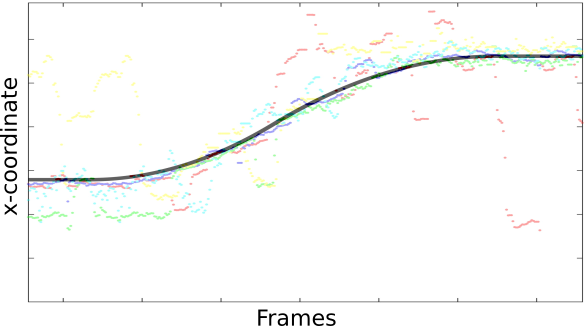}&
  \includegraphics[width=.3\textwidth]{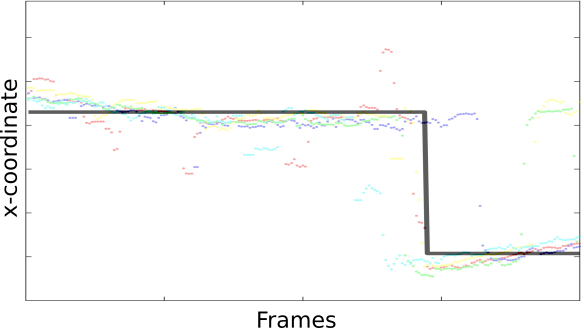}  \\
  {\footnotesize (a)} & {\footnotesize (b)} & {\footnotesize (c)} \\
  \end{tabular}
  \caption{The recorded eye gaze data from five users and the corresponding output (the optimal x position) of our algorithm. Results with three different type of gaze have been presented i.e. (a) fixation: when the user maintains the eye gaze around a location, our algorithm outputs a static segment; (b)  pursuit: a gradual movement of eye gaze results in a pan; (c) saccade: an abrupt transition in gaze leads to cut.} ~\label{fig:figure2}
\end{figure*}
\section{Related work}

Several media retargeting solutions have been proposed, and they can be broadly categorized into three different categories: discrete, continuous and cropping-based approaches. Discrete approaches~\cite{AS07,RubinsteinSA08,Shiftmap-09} judiciously remove and/or shift pixels to preserve salient media content. Examples include the relative shifting of pixels~\cite{Shiftmap-09}, or adding/removing connected seams (paths of low importance) from the image~\cite{AS07}. Continuous approaches~\cite{FeatureAware:2006} optimize a warping (mapping) from the source to the target media, with constraints designed to preserve salient media content resulting in a non-homogeneous transformation, with less important regions squeezed more than important ones. Some of these discrete and continuous approaches have been extended to video retargeting~\cite{RubinsteinSA08,WOLFGCO2007,KrahenbuhlLHG09}. However, discrete or continuous removal of pixels often results in visible artifacts such as squeezed shapes, broken lines or structures and temporal glitches/incoherence. 

A third retargeting approach selects a cropping window for each frame inside the original video. This approach eliminates visual artifacts, but some visual information is nevertheless lost in the cropping process. This is in spirit, the `pan and scan' process, where an expert manually selects a cropping window within a widescreen image to adapt the same to smaller aspect ratios. The goal is to preserve the important scene aspects, and the operator smoothly moves the cropping window as action shifts to a new frame position or introduces new cuts (for rapid transitions). 

Several efforts~\cite{LG06,DeselaersDN08,xiang2010video} have been made to automate the `pan and scan' or re-editing process. These approaches primarily rely on computational saliency (primarily bottom-up saliency based on spatial and motion cues) to discover important content, and then estimate a smooth camera path that preserves as much key content as possible. Liu and Gleicher~\cite{LG06} define the virtual camera path as a combination of piecewise linear and parabolic segments. Grundmann \etal~\cite{GrundmannKwatra2011} employ an $L(1)$ regularized optimization framework to produce stable camera movements thereby removing undesirable motion. {Gleicher and colleagues~\cite{GleicherM00,LG06,Heck07,gandhi:hal-01067093} relate the idea of a moving cropping window with virtual camera work, and show its application for editing lecture videos, re-editing movie sequences and multi-clip editing from a single sequence.} 

While human attention is influenced by bottom-up cues, it is also impacted by top-down cues relating to scene semantics such as faces, spoken dialogue, scene actions and emotions which are integral to the storyline~\cite{Subramanian2014,PET2015}. Leake~\etal~\cite{Leake2017} propose a computational video editing approach for dialogue-driven scenes which utilizes the script and multiple video recordings of the scene to select the optimal recording that best satisfies user preferences (such as emphasize a particular character, intensify emotional dialogues, \etc). {A more general effort was made by Galvane ~\etal ~\cite{galvane-aaai-10} for continuity editing in the 3D animated sequences. However, the movie script and the high-level cues used in these works may not always be available}. A number of other works have utilized \textit{eye tracking} data, which is indicative of the semantic and emotional scene aspects~\cite{katti2010making,Subramanian2014}, to infer salient scene content.   

Santella \etal~\cite{Santella-chi-06} employ eye tracking for photo cropping so as to satisfy any target size or aspect ratio. More recently, Jain~\etal~\cite{jain2014} perform video re-editing based on eye-tracking data, where RANSAC (random sampling consensus) is used to compute smooth B-splines that denote the cropping window path. Nevertheless, this approach is computationally very expensive as B-splines need to be estimated for every RANSAC trial. Also, the methodology involves implicit assumptions relating to the video content, and is susceptible to generating unmotivated camera motions due to imprecise spline blending.

\textbf{Contribution:} We propose a novel, optimization-based framework for video re-editing utilizing minimal user gaze data. To this end, it can work with any type of video (professionally created movies or wide-angle theatre recordings) of arbitrary length to produce a re-edited video of any given size or aspect ratio. Also, our optimization is $L(1)$ regularized, which economizes and smoothens virtual camera motion to mimic professional camera capture behavior, and ensures a smooth viewing experience. Also, since our methodology only requires approximate rather than accurate information regarding salient scene regions, we utilize eye-tracking data recorded with a low-end and affordable ($\approx$ 100\euro) eye tracker. We demonstrate that our method outperforms the state-of-the-art~\cite{jain2014} via experiments and a user study reflecting subjective human expressions concerning the quality of the re-edited video. 

\section{Problem Statement}
\label{sec:problem_statement}

%
%
The proposed algorithm takes as input (a) a sequence of frames $t=[1:N]$, where $N$ is the total number of frames; (b) the raw gaze points over multiple users, $g_t^i$, for each frame $t$ and subject $i$ and (c) the desired output aspect ratio. The output of the algorithm is the edited sequence to the desired aspect ratio, which is characterized by a cropping window parametrized by the $x$-position ($x^*_t$) and zoom ($z_t$) at each frame. The edited sequence introduces new panning movements and cuts within the original sequence, aiming to preserve the cinematic and contextual intent of the video. Before delving into the technical details, we put forward a discussion on desired characteristics of such an editing algorithm from a cinematographic perspective. We also discuss how the literature from cinematography~\cite{gramofshot,gramofedit,vidprodhand} inspires our algorithmic choices. 

Millerson and Owens~\cite{vidprodhand} stress on the aspect that the shot composition is strongly coupled with what viewers will look at. If viewers do not have any idea of what they are supposed to be looking at, they will look at whatever draws their attention (random pictures produce random thoughts). This motivates the choice of using gaze data in the re-editing process. Although, notable progress has been made in computationally predicting human gaze from images~\cite{mit-saliency-benchmark}, the cognitive gap still needs to be filled in~\cite{bylinskii2016should}. For this reason, we explicitly use the collected gaze data in the proposed algorithm as the measure of saliency. The algorithm then aims to align the cropping window with the collected gaze data.  


Importance of a steady camera has been highlighted in Thomson and Christopher's `Grammar of the shot'~\cite{gramofshot}. They suggest that (a) camera should not move without a sufficient motivation (as it may appear puzzling to the viewer) and brief erratic pans should be avoided and (b) a good pan/tilt movement should comprise of three components: a static period of the camera at the beginning, a smooth camera movement which "leads" the attention of the subject and a static period of the camera at the end. As discussed earlier, ideally we would like to center the cropping window around the most salient region (computed from gaze data), however we allow some relaxation using rectified linear unit function to avoid brief camera movements over small gaze variations. Furthermore, we use a combination of first, second and third order $L(1)$  norm regularization to achieve professional looking smooth pan/zoom (leading to piecewise static, linear and parabolic segments) ~\cite{GrundmannKwatra2011}. Figure~\ref{fig:figure2} illustrates the behavior of our algorithm with (a) gaze centered around a location where the algorithm outputs a perfectly static segment and (b) gradually varying gaze, where the algorithm outputs a smooth pan. 


The fast panning movements should be avoided, as it may lead to breakup of the movement~\cite{vidprodhand}. We impose constraints on the maximum panning speed in the optimization process, to make sure that fast panning movements do not occur. Apart from panning, another crucial parameter which needs to be controlled is amount of zoom, which is often correlated with the localization and intensity of the scenes~\cite{vidprodhand}. We use variance of gaze fixation data as the indicator of amount of zoom-in. If the gaze data is concentrated around a point for a long interval, it motivates zooming-in at that location and the high variance gaze suggests that a lot is going on in the scene and the camera should zoom out to include more content.  
Furthermore, to mimic human like behavior, we add a delay in zoom/pan movements (a computational camera starts panning at the instant an actor starts moving, while an actual cameraman takes a moment to respond to the action). 

Cuts are another form of transitions which are used when there is a need for quick change in impact. There should always be a reason to make the cut, the two shots before and after the cut should not be too similar (avoiding jump cut) and the pace of the cut should go along with the story (it should not be too fast and too slow)~\cite{gramofedit}. We argue that frames with  abrupt changes in gaze positions are good candidate locations for adding cuts, as that indicates the change in impact. To avoid jump cuts, we use a penalty term in the optimization process, quantifying the overlap in the compositions before and after the cut. Furthermore, we use a decaying exponential penalty to control the pace/rhythm of the cuts. Figure~\ref{fig:figure2}(c) illustrates an example where abrupt change in gaze location leads to a cut.

\section{Method}

Our algorithm consists of two steps.{The first step uses dynamic programming to detect a path ($\epsilon = \{r_t\}_{t=1:N}$) for the cropping window which maximizes the amount of gaze and time stamps appropriate to introduce new cuts which are cinematically plausible.} The second step optimizes over the path $(\{\text{r}_t\})$ to mimic the professional cameraman behavior, using a convex optimization framework (converting the path into piecewise linear, static and parabolic segments, while accounting for the original cuts in the sequence and the newly introduced ones). We first explain the data collection process and then describe the two stages of our algorithm.  
%
\subsection{Data collection}
\label{sec:data_cllection}
We selected a variety of clips from movies and live theatre. A total of 12 sequences are selected from four different feature films and cover diverse scenarios like dyadic conversations, conversations in crowd, action scenes, \etc~The clips include a variety of shots such as close ups, distant wide angle shots, stationary and moving camera \etc~ The native aspect ratio of all these sequences is either 2.76:1 or 2.35:1. The pace of the movie sequences vary from a cut every 1.6 seconds to no-cuts at all in a 3 minute sequence. The live theatre sequence were recorded from dress rehearsals of Arthur Miller's play `Death of a salesman' and Tennessee Williams' play `Cat on a hot tin roof'. All the 4 selected theatre sequences were recorded from a static wide angle camera covering the entire stage and have an aspect ratio of 4:1. These are continuous recordings without any cuts. The combined set of movie and live theatre sequences amount to a duration of about 52 minutes (minimum length of about 45 seconds and maximum length of about 6 minutes).

Five naive participants with normal vision (with or without lenses) were recruited from student community for collecting the gaze data. The participants were asked to watch the sequences resized to a frame size of 1366 $\times$ 768 on a 16 inch screen. The original aspect ratio was preserved during the resizing operation using letterboxing. The participants sat at approximately 60 cm from the screen. Ergonomic settings were adjusted prior to the experiment and system was calibrated. PsychToolbox ~\cite{kleiner2007s} extensions for MATLAB were used to display the sequences. The sequences were presented in a fixed order for all participants. The gaze data was recorded using the 60 Hz Tobii Eyex, which is an easy to operate, low cost eye-tracker.  
\subsection{Gaze as an indicator of importance}
The basic idea of video retargeting is to preserve what is important in video by removing what is not. We explicitly use gaze as the measure of importance and propose a dynamic programming optimization, which takes as input the gaze tracking data from multiple users and outputs a cropping window path which encompasses maximal gaze information. The algorithm also outputs the time stamps to introduce new cuts (if required) for more efficient storytelling. Whenever there is an abrupt shift in the gaze location, introducing a new cut in the cropping window path is a preferable option over panning movement (as fast panning would appear jarring to the viewer). However, the algorithm penalizes jump cuts (ensuring that the cropping window locations, before and after the cut are distinct enough) as well as many cuts in short succession (it is important to give the user sufficient time to absorb the details before making the next cut).

%
%
%
\begin{figure*}[h!]
  \centering
  \includegraphics[width=0.9\textwidth]{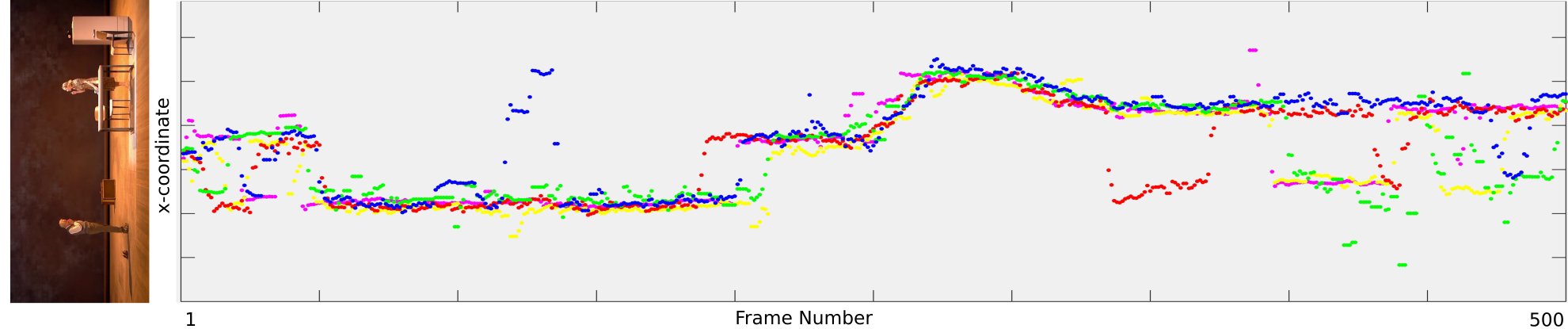}
  \includegraphics[width=0.9\textwidth]{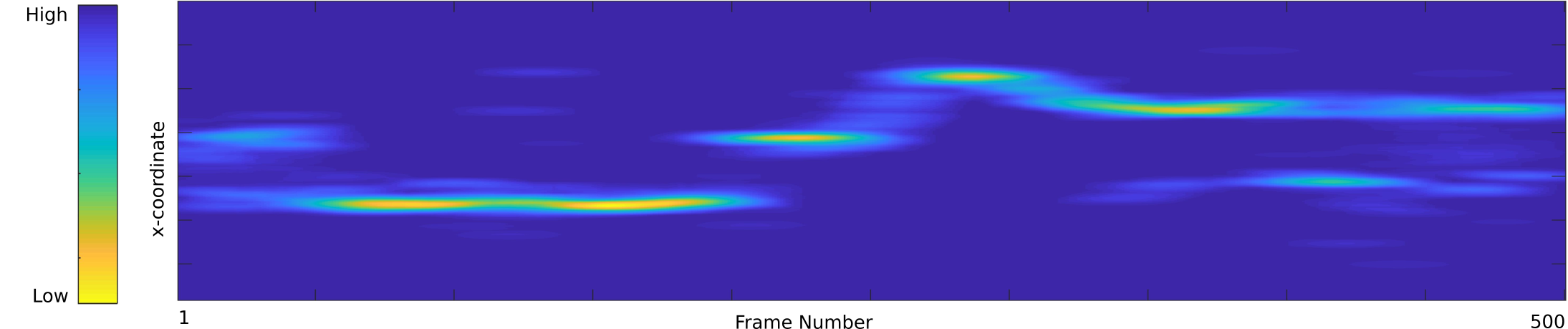}
  \includegraphics[width=0.9\textwidth]{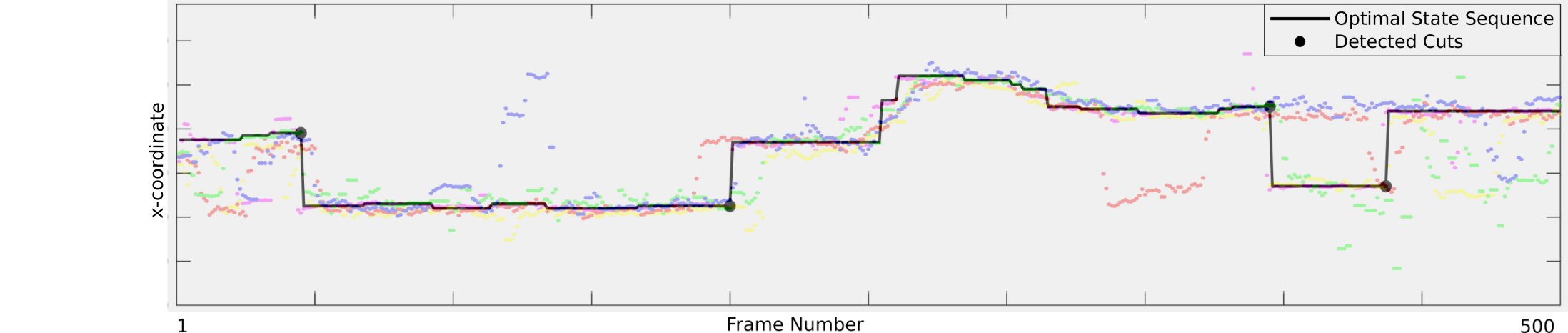}
  \caption{The x-coordinate of the recorded gaze data of 5 users for the sequence "Death of a Salesman" (top row). Gaussian filtered gaze matrix over all users, used as an input for optimization algorithm (middle row). The output of optimization: The optimal state sequence (black line) along with the detected cuts (black circles) for an output aspect ratio of 1:1(bottom row).}~\label{fig:figure3}
\end{figure*}
%

More formally, the algorithm takes as input the raw gaze data $\text{g}^i_t$ of $\text{i}^{th}$ 
user for all frames $t = [1 : N]$ and outputs a state $\epsilon = \{\text{r}_t\}$ for each frame. Where the state $r_t  \in [1:W_o] $ (where $W_o$ is width of the original video frames) selects one among all the possible cropping window positions. The optimization problem aims to minimize the following cost function:
%
%
\begin{equation} 
\label{eqn:global_optimization}
E(\epsilon) = \sum_{t=1}^N E_s(r_t) + \lambda \sum_{t=2}^N E_t(r_{t-1},r_t,d)
\end{equation}
where the first term $E_s$ penalizes the deviation of $r_t$ from the gaze data at each frame and $E_t$ is the transition cost, which computes the cost of transition from one state of another (considering both the options of camera movement and cut). Given the raw gaze data $g^i_t$, the unary term $E_s$ is defined as follows:.
$$
\begin{aligned}
& & & E_s(r_t) = M_s(r_t,t) &\\
& \text{where} & & M_s(x,t) = \left(\sum_{i=1}^u M^i(x,t) \right) * \mathcal{N}(0,\,\sigma^{2}) & \\
& & & M^i(x,t) = \begin{cases}
    -1  & \text{if } x = g^i_t\\
    0  & \text{otherwise} \\
    \end{cases} &
\end{aligned}
$$
Here, $ M^i(x,t) $ is a $W_o \times N$ matrix of $i^{th}$ user gaze data and  $M_s(x,t)$ is the sum of gaussian filtered gaze data over all users. Figure~\ref{fig:figure3} (middle row) shows an example of matrix $M_s$ computed from the corresponding gaze data (top). Essentially $E_s(r_t)$ is low, if $r_t$ is close to the gaze data and increases as $r_t$ deviates from the gaze points.  Using the combination of multiple user gaze data, makes the algorithm robust to both the noise induced by the eye-tracker and the momentary erratic eye movements of some users. {We use a Gaussian filter (with standard deviation $\sigma$) over the raw gaze data to better capture the overlap between multi-user data, giving a lowest unary cost to areas where most users look at.}
%
%

The pair-wise cost $E_t $ considers a case wise penalty. The penalty differs if  there is a new cut introduced or not. If there is no cut introduced, it is desired that the the new state be closer to the previous state. If a new cut is introduced, it is desired to avoid a jump cut and also leave sufficient time from the previous cut. The term $E_t$ is defined as follows:
%
\begin{equation}
E_t(r_{t-1},r_t, d) =
\begin{cases}
    \left(1- e^{\frac{-4|r_t-r_{t-1}|}{ W } } \right) &  |r_t-r_{t-1}| \leq W\\
    \left( 1 + e^{\frac{-|d|}{D}} \right) & |r_t-r_{t-1}| > W ,
\end{cases}
\label{eqn:cutdp}
\end{equation} 
where $d$ is the duration from the previous cut and $D$ is a parameter which controls the cutting rhythm and {can be tuned for faster or slower pace of the scene. We set the value of $D$ to 200 frames, which is roughly the average shot length used in movies between the 1983 and 2013~\cite{cutting2015shot}}. The first case in Equation \ref{eqn:cutdp} is considered, when the difference in consecutive states is less than $W$, {the minimum width to avoid jump cut (we assume occurrence of jump cut if overlap between consecutive cropping windows is more than 25\%}). The cost is $0$ when $r_t = r_{t-1}$ and gradually saturates towards $1$, when $|r_t-r_{t-1}|$ approaches $W$. A transition of more than $W$ indicates possibility of a cut, and then the pairwise cost is driven by the duration from the previous cut. The cost gradually decreases with increase in duration from the previous cut.

Finally, we solve Equation~\ref{eqn:global_optimization} using Dynamic Programming (DP). {The algorithm selects a state $r_t$ for each time $t$ from the given possibilities ($W_o$ in this case). We build a cost matrix $C(r_t,t)$ (where $r_t \in  [1:W_o] $ and $t \in [1:N]$ ). Each cell in this table is called a $node$. The recurrence relation used to construct the DP cost matrix is a result of the above energy function and is as follows:
$$
\begin{aligned}
& & C(r_t,t) = \begin{cases}
E_s(r_t) & t=1 \\
\underset{r_{t-1}}{\text{min}} \left[ E_s(r_t) + \lambda *E_t(r_{t-1},r_t,d) +C(r_{t-1},t-1)\right] & t>1
\end{cases}
\end{aligned}
$$ For each node $(r_t, t)$ we compute and store the minimum cost $C(r_t,t)$ to reach it. A cut $c_t$ is introduced at frame $t$, if the accumulated cost is lower for introducing a cut than keeping the position constant or panning the window. Backtracking is then performed from the minimum cost node in the last column to retrieve the desired path. Finally, the output of the algoritm is the optimized cropping window path $\epsilon = \{\text{r}_t\}$ and the set of newly introduced cuts $\{ c_t \}$. The time complexity of the algorithm is $O({W_o}^2N)$ and the space complexity is $O(W_oN)$, which are both linear with $N$.} An example of the generated optimization result is illustrated in Figure~\ref{fig:figure3} (bottom row).


\subsection{Optimization for cropping window sequence}
The output of the dynamic programming optimization gives a cropping window path which maximizes the inclusion of gaze data inside the cropping window and the location of the cuts. However, this cropping window path does not comply with cinematic principles (leading to small erratic and incoherent movements). We further employ an L(1) regularized convex optimization framework, which aims to convert the rough camera position estimates into smooth professional looking camera trajectories, while accounting for cuts (original and newly introduced ones) and other relevant cinematographic aspects, as discussed in Section~\ref{sec:problem_statement}. This optimization takes as input, the original gaze data; the initial path estimate ($\epsilon = \{r_t\}_{t=1:N}$ ); the original cuts in the sequence, if any (computed using~\cite{apostolidis2014fast}); the newly introduced cuts $c_t$ and outputs the optimized virtual camera trajectory, $\xi = \{(x^*_t,z_t)\}_{t=1:N}$. The optimization consists of several cost terms and constraints and we describe each of them in detail:



\subsubsection{Data term:} The data term penalizes deviation of the virtual camera path (cropping window path) from the initial estimates (which eventually is capturing the gaze behavior). The term is defined as follows:
\begin{equation}
 D(\xi) = \sum^N_{t=1} \left( max\left[|x^*_t - r_t|-\tau,0 \right] \right)^2
\end{equation}
The function penalizes if the optimized sequence $x^*_t$ deviates from the initial estimate of the camera position $r_t$. However, it is relaxed with a rectifier linear unit function to avoid the penalty for small gaze movements and in turn to avoid brief and erratic camera movements. To summarize, the above cost function incurs a penalty only if the optimal path $x^*_t$ varies from $r_t$, with more than a threshold, $\tau$.
\subsubsection{Movement regularization}
As discussed in Section~\ref{sec:problem_statement}, smooth and steady camera movement is necessary for pleasant viewing experience~\cite{gramofshot}. Professional cameramen avoid unmotivated movements and keep the camera as static as possible. When the camera is moved, it should start with a segment of constant acceleration followed by a segment of constant velocity and should come to a static state with a segment of constant deceleration. Early attempts modeled this behavior with heuristics~\cite{GL08}, however recent work by Grundmann ~\etal ~\cite{GrundmannKwatra2011} showed that such motions could be computed as the minima of an $L(1)$ optimization. In the similar spirit, we introduce three different penalty terms to obtain the desired camera behavior. 

When $L(1)$ norm term is added to the objective to be minimized, or constrained, the solution typically has the argument of the $L(1)$ norm term sparse (i.e., with many exactly zero elements). The first term, penalizes the $L(1)$ norm over the first order derivative, inducing static camera segments:
\begin{equation}
M_1(\xi) = \sum_{t=1}^{N-1} ( |x_{t+1}^*-x_t^*|).
\end{equation}
The second term induces constant velocity segments by minimizing accelerations:
\begin{equation}
\begin{split}
M_2(\xi) = \sum_{t=1}^{N-2} ( |x_{t+2}^*-2x_{t+1}^* + x_{t}^*|).
\end{split}
\end{equation}
The third term minimizes jerk, leading to segments of constant acceleration:
\begin{equation}
\begin{split}
M_3(\xi) = \sum_{t=1}^{N-3} ( |x_{t+3}^*-3x_{t+2}^* + 3x_{t+1}^* - 3x_{t}^*|).
\end{split}
\end{equation}
Combining these three penalties yields camera movements consisting of distinct static, linear and parabolic segments.

\subsubsection{Zoom}

We perform zoom by varying the size of the cropping window (decreasing the size of the cropping window results in a zoom-in operation, as it makes the scene look bigger). The amount of zoom is decided based on the standard deviation of the gaze data, taking inspiration from previous work on gaze driven editing~\cite{jain2014}. However, we use gaze fixations instead of the raw gaze data for computing the standard deviation. We observed that using fixation gives added robustness over the outliers/momentary noise. We use the EyeMMV toolbox~\cite{krassanakis2014eyemmv} for computing the fixation, with a duration of 200 ms. 

\begin{figure*}[h!]
  \centering
  \includegraphics[width=\textwidth]{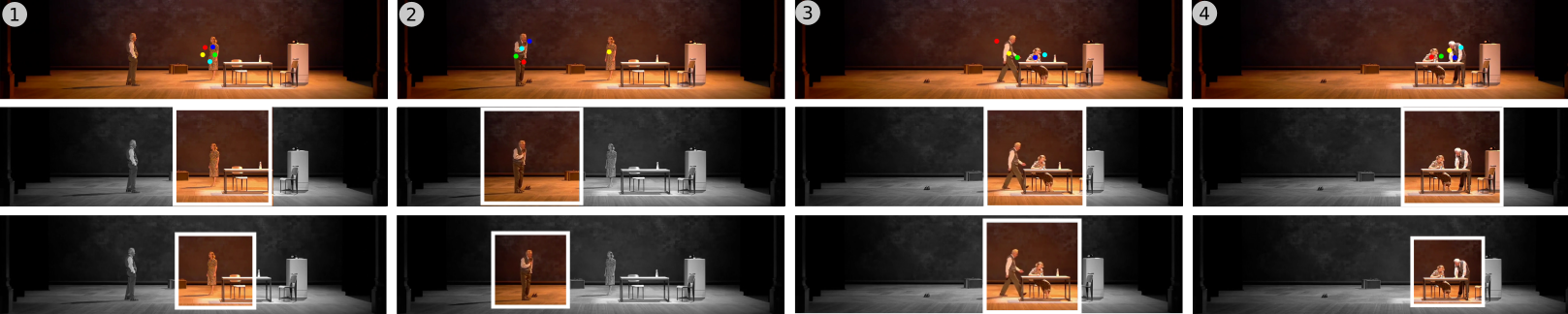} 
    \vspace*{-0.9em} \par \hfill \fontsize{5}{4}\selectfont{\copyright Celestin, Theatre de Lyon }\\
    \vspace*{0.3em}
  \caption{An example sequence where our method performs zoom-in and zooms-out action. The original sequence with overlaid gaze data (Top). The output of our algorithm without zoom (Middle) and with zoom (Bottom).}~\label{fig:zoom}
\end{figure*}

Let $\sigma_t$ be the standard deviation of the gaze data at the fixation points at each frame. The ratio 
\begin{equation}
\rho_t = 1-0.3*\left(1-\frac{\sigma_t}{\underset{k}{max}(\sigma_k)}\right)
\end{equation}
$\rho_t  \in [0.7, 1]$ is used as an indicator for the amount of zoom at each frame ($\rho_t =1$ corresponds to the largest feasible cropping window and zoom-in happens as value of $\rho_t$ decreases). We add the following penalty terms in the optimization:
\begin{equation}
 Z(\xi) = \sum^N_{t=1} \left(z_t - \rho_t\right)^2,
\end{equation}
which penalizes the deviation of zoom from the value $\sigma_t$ at each frame. 
We further add  $L(1)$ regularization terms (first order ($M^z_1$), second order ($M^z_2$) and third order ($M^z_3$) ) over $z_t$ to the objective function to avoid small erratic zoom-in zoom-out movements and ensure that whenever zoom action takes place, it occurs in a smooth manner.  

\subsubsection{Inclusion and panning constraints:}

We introduce two sets of hard constraints. The first constraint enforces the cropping window to be always within the original frame i.e. $ \frac{W_r}{2} < x^*_t < W_o - \frac{W_r}{2} $, which ensures a feasible solution (where $W_r$ is the width of the retarget video). We add second constraint as upper bound on the velocity of the panning movement, to avoid fast panning movements. Cinematographic literature~\cite{vidprodhand} suggests that a camera should pan in such a way that it takes an object at least 5 seconds to travel from one edge of the screen to the other. This comes out to roughly 6 pixels/frame for the video resolution used in our experiments and leads to following constraint: $ |x^*_t - x^*_{t-1}| \leq 6$. 

\subsubsection{Accounting for cuts : original and newly introduced}

Our algorithm is agnostic to the length and type of the video content; this means that the original video may include arbitrary number of cuts (original and newly introduced). This is in contrast to previous approaches~\cite{jain2014}, which solve the problem on a shot-by-shot basis. This generalization is achieved by relaxing the movement regularization around cuts. The following two properties are desired in the periphery of a cut: (a) The transition at the cut should be sudden; and (b) The camera trajectory should be static just before and after the cut, as cutting with moving cameras can cause the problem of motion mismatch~\cite{gramofedit};

To induce sudden transition, we make all penalty terms zero at the point of cut. We also make the data term zero, $p$ frames before and after every cut to account for the delay a user takes to move from a gaze location to another (although the change of focus in the scene is instantaneous, in reality the viewer takes a few milliseconds to shift his gaze to the new part of the screen). Similarly to induce static segments before and after the cut, we make the second and third order $L(1)$ regularization zero in the same interval. However, we keep the first order $L(1)$ regularization term non zero on the entire optimization space, except at the exact point to cut, to allow for the transition.{ The parameter $p$ is set to 5, because we use third order $L(1)$ term which uses 4 previous values.} 

\subsubsection{Energy Minimization:}
Finally, the problem of finding the optimal cropping window sequence can be simply stated as a problem of minimizing a convex cost function with linear constraints. The overall optimization function is defined as follows: 
\begin{equation}
\begin{aligned}
& \underset{x^*, z}{\text{minimize}}
& & D(\xi)  + \lambda_1 M_1(\xi)  + \lambda_2 M_2(\xi) + \lambda_3 M_3(\xi) + \\
& & & Z(\xi)  + \lambda_1 M^z_1(\xi)  + \lambda_2 M^z_2(\xi) + \lambda_3 M^z_3(\xi)  \\
& \text{subject to}
& & \frac{W_r}{2} \leq x^*_t \leq W_o - \frac{W_r}{2} \\
& & & |x^*_t - x^*_{t-1}| \leq 6, \\
& & & 0.7 \leq z_t \leq 1, \; t = 1, \ldots, N-1.  \\ 
\end{aligned}
\end{equation}
As discussed in Section~\ref{sec:problem_statement}, the optimal cropping window, $x^*_t$ usually starts panning, at the instant the actor starts moving, while the actual cameraman takes a moment to respond for the action. To account for this, we delay the optimal cropping window path, $x^*_t$, by 10 frames for each shot.{ The parameters, $\lambda_1$, $\lambda_2$, $\lambda_3$ can be changed to vary the motion model. Currently, we keep $\lambda_1$ higher, preferring static segments.
}


\section{Results}
The results are computed on all the 12 clips (8 movie sequences \& 4 theatre sequences) mentioned in Section~\ref{sec:data_cllection}. All the sequences were retargeted from their native aspect ratio to 4:3 and 1:1 using our algorithm. We also compute results using Gaze Driven Editing (GDR)~\cite{jain2014} algorithm by Jain~\etal $ $ for the case of 4:3 aspect ratio, over all the sequences. The results with GDR were computed by first detecting original cuts in the sequences and then applying the algorithm shot by shot. Some of the example results and comparisons are shown in Figure~\ref{fig:zoom}, Figure~\ref{fig:eakta1} and Figure~\ref{fig:eakta2}. An explicit comparison on output with and without zoom is shown in  Figure~\ref{fig:zoom}. All the original and retargeted sequences are provided in the supplementary material. 
%

We used CVX~\cite{cvx} toolbox with MOSEK ~\cite{mosek} for convex optimization. The parameters used for the algorithm are given in Table~\ref{parameters}. Same set of parameters are used for all theatre and movie sequences.
%
%
%
\begin{table}[h!]
\fontsize{7}{7}\selectfont
\renewcommand{\arraystretch}{1.2}
\centering
\begin{tabular}{|c|c|c|c|c|c|c|c|c|}
\hline
\textbf{Parameter} & {$\lambda$} & $\sigma$ & D   & {$W$} &$\lambda_1$ & $\lambda_2$ & $\lambda_3$ & {$\tau$} \\ \hline
\textbf{Values}  & {2} & 15 & 200 & {0.75$W_{r}$} & 5000 & 500 & 3000 & {0.1$W_r$} \\ \hline
\end{tabular}
\vspace{-.1cm}
\caption{\label{parameters}Parameters for path optimization algorithm {and }cut detection algorithm.}
%
\vspace{0.2cm}
\fontsize{8}{8}\selectfont
\renewcommand{\arraystretch}{1.2}
\centering
\begin{tabular}{|c|c|c|}
\hline
\textbf{Type} & \textbf{Our Method} & \textbf{GDR}  \\ \hline
\textbf{All} & $89.63\%(\sigma = 4.96)$ & $76.26\%(\sigma = 9.71)$ \\
\textbf{Movies} & $90.28\%(\sigma = 4.63)$ & $77.76\%(\sigma = 10.3)$  \\
\textbf{Theatre} & $85.95\%(\sigma = 4.43)$ & $74.02\%(\sigma = 6.27)$\\ \hline
\end{tabular}
\caption{\label{gaze_included} Comparing our method with GDR based on the mean percentage of gaze data included within the retargeted video at 4:3 aspect ratio.}
\label{gaze_include}
\vspace{-.3cm}
\end{table}
\subsection{Runtime}
The proposed algorithm optimizes the cropped window for any length of video sequence with arbitrary number of cuts. Hence, it is independent of the video resolution and number of shots/cuts. The proposed algorithm takes around 40 sec for a 6 min video sequence (processing around 220 frames per second) on a laptop with i3 processor and 4GB RAM whereas~\cite{jain2014} takes around 40 min for 30 sec sequence. We empirically observed that the complexity of our algorithm increases linearly with number of frames and takes about 10 minutes for retargeting a 90 minute movie.
%
\subsection{Included gaze data}
One measure for evaluating retargeting performance is to compute the percentage of gaze data included inside the cropped window, as suggested in ~\cite{chamaret2008attention} and ~\cite{jain2014}. Table ~\ref{gaze_include} shows the average percentage of gaze data included over all the retargeted videos at 4:3 aspect ratio with our method and GDR. The global perspective and flexibility of our method allows it to capture considerably more gaze data than GDR. On average, our method is able to include about 13\% more gaze data and the numbers reflect for both movie and theatre sequences. Smaller deviation ($\sigma = 4.96$ for our method, compared to $\sigma = 9.71$ for GDR) across sequences also confirms content agnosticism of our algorithm. The proportion of included gaze reduces to about 81.8\% when the retargeted videos are rendered at 1:1 aspect ratio. In other words, when retargeting a video from 2.76:1 to 1:1, our algorithm preserves around 81\% of gaze data while losing around 63\% of the screen space.


\begin{figure*}[h!]
  \centering
\includegraphics[width=\textwidth]{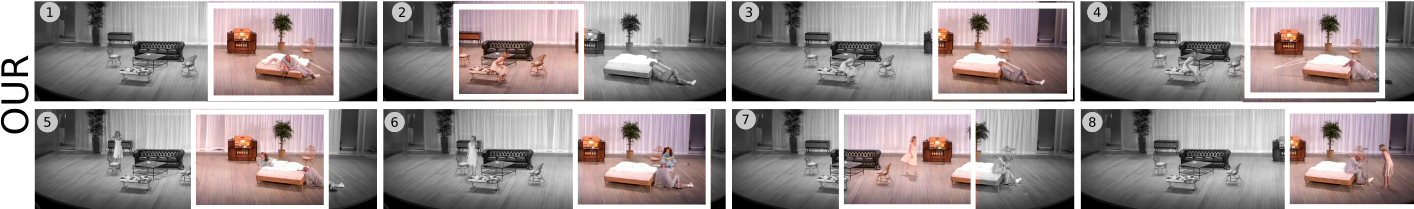}
\includegraphics[width=\textwidth]{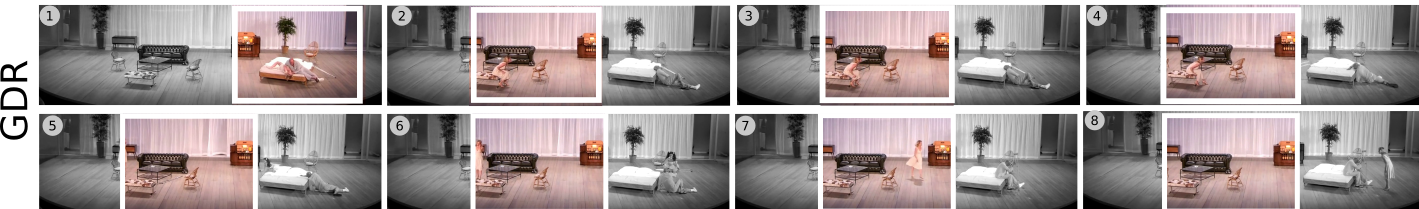}
   \vspace*{-1em} \par \hfill \fontsize{5}{4}\selectfont{\copyright Celestin, Theatre de Lyon} \\
    \vspace*{0.4em}
  \break
 \includegraphics[width=\textwidth]{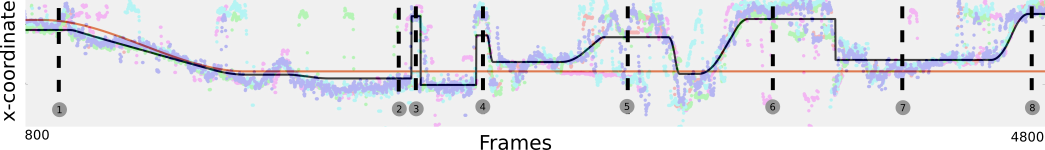}
 \caption{The figure shows original frames and overlaid outputs from our method and GDR (coloured, within white rectangles) on a long sequence. Plot shows the $x$-position of the center of the cropping windows for our method (black curve) and GDR (red curve) over time. Gaze data of 5 users for the sequence are overlaid on the plot. Unlike GDR, our method does not involve hard assumptions and is able to better include gaze data (best viewed under zoom).}~\label{fig:eakta1}
 \vspace{-0.6cm}
\end{figure*}
\begin{figure}[b]
  \centering
   \vspace{-0.6cm}
\includegraphics[width=\columnwidth]{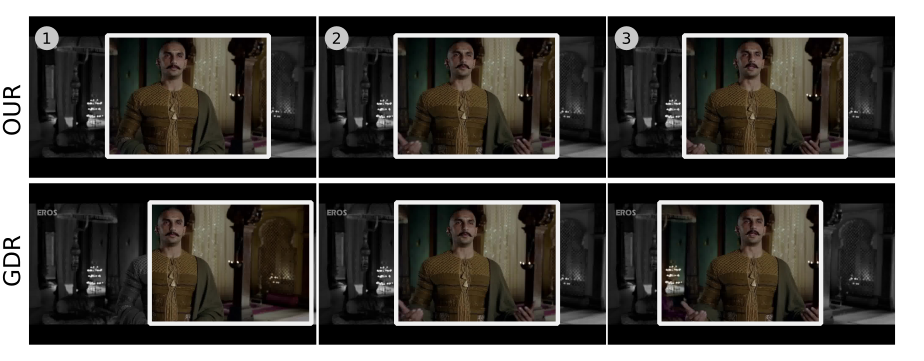}
\vspace*{-0.6em} \fontsize{5}{4}\selectfont{\par \hfill \copyright Eros International} \\    \vspace*{0.4em}
  \break
 \includegraphics[width=\columnwidth]{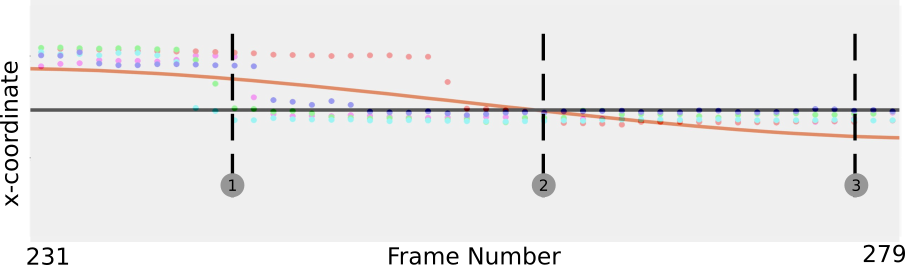}
 \caption{Frames from the original sequence cropped by the output of our algorithm and GDR (white rectangles). Corresponding gaze data (below) reveals that gaze is broadly cluttered around the shown character. Our algorithm produces a perfectly static virtual camera segment (black curve), while GDR results in unmotivated camera movement (red curve).}~\label{fig:eakta2}
  \vspace{-0.2cm}
\end{figure}
%
\subsection{Qualitative evaluation}
We compare our results with GDR in Figures~\ref{fig:eakta1} and~\ref{fig:eakta2}. The hard assumptions of at most two pans and a single cut in a shot limits the applicability of GDR on longer sequences, as seen in Figure~\ref{fig:eakta1}. The cropping window becomes constant after frame 3 and misses the action taking place in the scene.  In contrast, our method consistently follows the action by using smooth pans (with smooth interpolation between static segments) and introducing new cuts. GDR can also lead to sudden unmotivated camera movements as it is applied on individual shots. An example is shown in Figure~\ref{fig:eakta2}, where gaze is primarily fixated on the main character and the GDR algorithm produces an unnecessary pan. Conversely, our method produces a perfectly static camera behavior via retargeting considering the \textit{entire video}. Results over all sequences achieved using GDR and our method are provided in the supplementary material for further qualitative comparisons. 

\section{User study evaluation}
\begin{figure*}[htbp]
\includegraphics[width=0.32\linewidth]{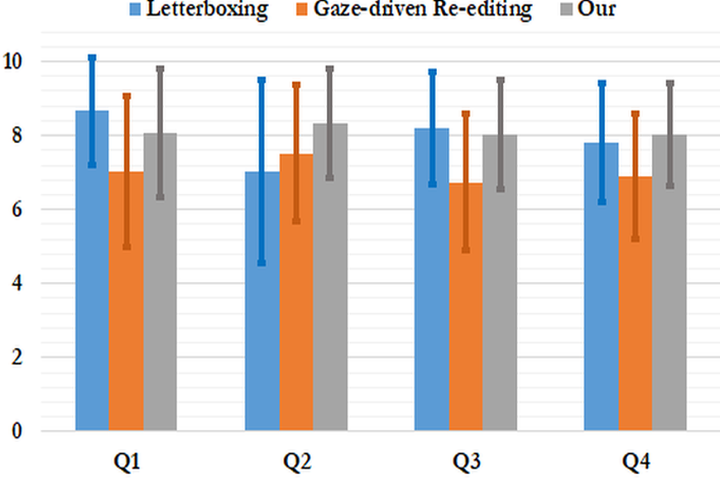}\hspace{0.1cm}\includegraphics[width=0.32\linewidth]{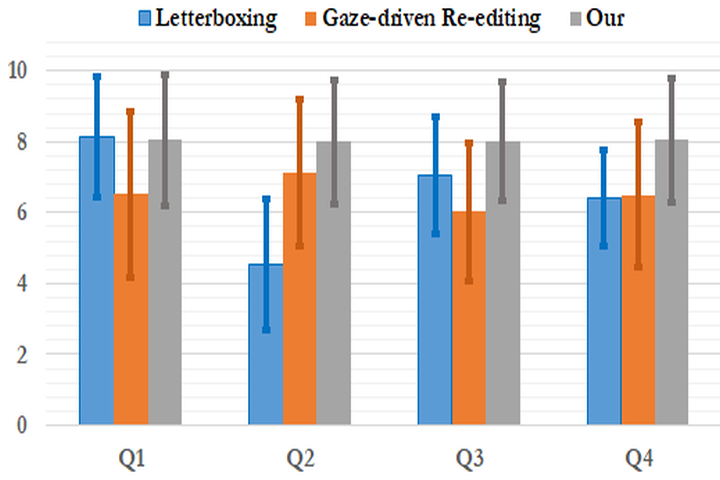}\hspace{0.1cm}\includegraphics[width=0.32\linewidth]{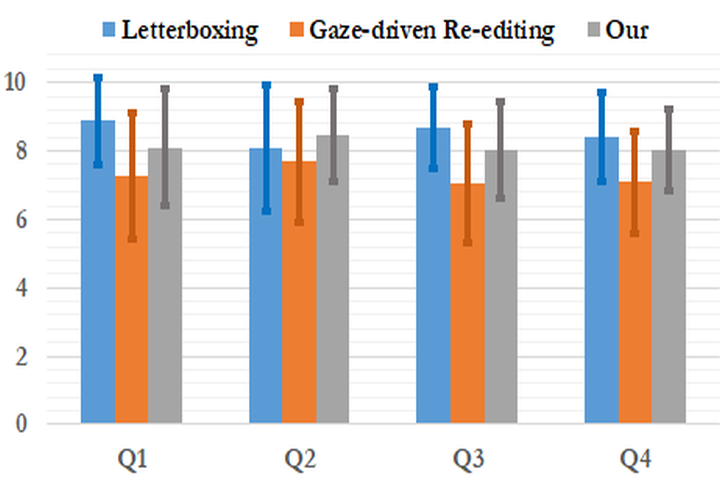}
\caption{\label{us_graphs}Bar plots showing scores provided by users in response to the four questions (Q1-Q4) posed for (left) \textit{all} clips, (middle) \textit{theatre} clips and (right) \textit{movie} clips. Error bars denote unit standard deviation.}
\vspace{-.2cm}
\end{figure*}

The primary motivation behind employing an $L(1)$ regularized optimization framework is to produce a smooth viewing experience. In order to examine whether our gaze-based retargeting approach positively impacted viewing experience, we performed a study with 16 users who viewed 20 original and re-edited snippets from the 12 videos used in our study. The re-edited snippets were generated via (a) letterboxing, (b) the gaze driven re-editing (GDR) of Jain~\etal~\cite{jain2014} and (c) our approach. Details of the user study are presented below. The output aspect ratio was 4:3 in both the cases of GDR and ours, without considering the zoom in both cases. 

\subsection{Materials and methods}
16 viewers (22--29 years of age) viewed 20 snippets of around 40 seconds length initially in their original form, followed by re-edited versions produced with letterboxing, GDR and our method shown in random order. Similar to the eye-tracking study, viewers watched the videos on a 16 inch screen at 1366 $\times$ 768 pixel resolution from around 60 cm distance. We followed a randomized 4$\times$4 Latin square design so that each viewer saw a total of five snippets (and each of them in four different forms), and such that all 20 snippets were cumulatively viewed by the 16 viewers.  

We slightly modified the evaluation protocol of Jain~\etal~\cite{jain2014}, who simply asked viewers regarding their preferred version of the original clip at a reduced aspect ratio. We instead asked viewers to provide us with real-valued ratings on a scale of 0--10 in response to the following questions:

\begin{itemize}
\item[$Q1.$]\hspace{0.5em} Rate the edited video for how effectively it conveyed the scene content with respect to the original (scene content effectiveness or SCE).
\item[$Q2.$]\hspace{0.5em} Rate the edited video for how well it enabled viewing of actors' facial expressions (FE).
\item[$Q3.$]\hspace{0.5em} Rate the edited video for how well it enabled viewing of scene actions (SA).
\item[$Q4.$]\hspace{0.5em} Rate the edited video for viewing experience (VE).
\end{itemize}   
These four questions were designed to examine how faithfully the examined video re-editing methods achieve the objective of the `pan and scan' approach. $Q1$ relates to how well the re-editing process preserves the scene happenings and semantics. $Q2$ and $Q3$ relate to the cropping window movements, and how well they capture the main scene actors and their interactions (\eg, when feasible, it would be preferable to watch both boxers in a boxing match rather than only the one who is attacking or defending). $Q4$ was added to especially compare the smoothness of the cropping window trajectory in the GDR and our approaches, and with the larger objective that re-editing approaches should not only capture salient scene content but should also be pleasing to the viewer's eyes by enforcing (virtual) camera pan and zoom only sparingly and when absolutely necessary. 

Of the 20 snippets, 14 were extracted from movie scenes, and the remaining from theatre recordings. Since the content of these scenes varied significantly (professionally edited vs wide angle static camera recorded), we hypothesized that the re-editing schemes should work differently, and have different effects on the two video types. Overall, the study employed a 3$\times$2 within-subject design involving the \textit{re-editing} technique (letterboxing, GDR or our) and the \textit{content type} (movie or theatre video) as factors.

\subsection{User data analysis}     
Figure~\ref{us_graphs} presents the distribution of user scores in response to questions $Q1-Q4$ for \textit{all}, \textit{theatre} and \textit{movie} clips. In order to examine the impact of the snippet content and re-editing techniques, we performed a 2-way unbalanced ANOVA (given the different number of movie and theatre snippets) on scores obtained for each question. 

For $Q1$, Figure~\ref{us_graphs} (left) shows that letterboxing scores are highest followed by our method and GDR respectively. This is to be expected as letterboxing preserves all the scene content with only a loss of detail, while re-editing techniques are constrained to render only a portion of the scene that is important. ANOVA revealed the main effect of both editing technique and content type. A post-hoc Tukey test further showed that the SCE scores were significantly different for Letterboxing (mean SCE score = 8.7) and GDR (mean SCE score = 7) at $p$<0.0001, as well as our method (mean SCE score = 8.1) and GDR at $p$<0.001, while the scores for our method and letterboxing did not differ significantly. These results suggest that \textit{\textbf{our retargeting method preserves scene content better than GDR, but only slightly worse than letterboxing}}.

\begin{table}[htbp]
\fontsize{9}{9}\selectfont
\renewcommand{\arraystretch}{1.2}
\centering
\begin{tabular}{|c|ccc|}
\hline
\textbf{Type} & \textbf{Our} & \textbf{Letterboxing} & \textbf{GDR} \\ \hline
\textbf{All} & 43.6 & 39.3 & 17.1 \\
\textbf{Movies} & 35.9 & 46.6 & 17.5 \\
\textbf{Theatre} & 65.2 & 18.8 & 16\\ \hline
\end{tabular}
\caption{\label{us_table}User preferences (denoted in \%) based on averaged and $z$-score normalized user responses for questions Q1-Q4.}
\vspace{-.6cm}
\end{table}

Scores for $Q2$ in Figure~\ref{us_graphs} (left) show that our method performs better than the two competitors. {ANOVA revealed the significant effects of the content type ($p$<0.05) and re-editing technique ($p$<0.0001) in this case}. A Tukey test showed that the FE scores for our method (mean FE score = 8.3) were significantly different from either GDR (mean FE score = 7.5) or letterboxing (mean FE score = 7). These result reveal that \textbf{\textit{our retargeting reveals actors' expressions most effectively, while letterboxing performs worst in this respect due to loss of scene detail}}. For scene actions however, letterboxing again scored the highest while GDR scored the lowest. Tukey test for SA showed that both letterboxing (mean SA score = 8.2) and our approach (mean SA score = 8) performed significantly better than GDR {(mean SA score = 7.2)}, while the difference between letterboxing and our approach was insignificant. \textit{\textbf{So, our retargeting performs only slightly worse than letterboxing with respect to preserving scene actions}}. Finally, \textbf{\textit{our method and letterboxing achieve very comparable scores for viewing experience}}, with both receiving significantly higher scores (mean VE score $\approx$ 8) than GDR (mean VE score = 6.9).


Figures~\ref{us_graphs} (middle) and (right) show the user score bar plots corresponding to theatre and movie snippets. Quite evidently, our FE, SA and VE scores are the highest for theatre clips, and are found to be significantly higher than either GDR or letterboxing via Tukey tests. Nevertheless, the superiority of our method diminishes for movie clips, with letterboxing and our approach performing very comparably in this case. Except for FE with theatre videos, GDR scores the least for all other conditions. These results again show that our method is able to capture theatre scenes best, and the main difference between theatre and movie scenes is that action is typically localized to one part of the stage in theatre scenes, while directors tend to effectively utilize the entire scene space in their narrative for movie scenes. Since our method is inherently designed to lose some scene information due to the use of a cropping window, it generates an output comparable to letterboxing in most cases. However, GDR performs the worst among the three considered methods primarily due to unmotivated camera movements and heuristic motion modeling which fails on longer shots. Table~\ref{us_table} tabulates the percentage of viewers that preferred our method over letterboxing and GDR upon $z$-score normalizing and averaging responses for {$Q1-Q4$}. The numbers again reveal that our method is most preferred for theatre, but loses out to letterboxing for movie clips. Cumulatively, \textbf{\textit{subjective viewing preferences substantially favor our method as compared to gaze based re-editing}}.

\section{Discussion}
We present an algorithm to retarget video sequences to smaller aspect ratios based on the gaze tracking data collected over multiple users. The algorithm uses a two stage optimization procedure to preserve the gaze data as much a possible while adhering to cinematic principles concerning pans, zooms and cuts. In contrast to previous approaches~\cite{jain2014}, we employ a heuristic-free motion modeling and our algorithm does not make any assumptions on the input sequences (both in terms of type and length). This makes our algorithm applicable for re-editing existing movie sequences as well as edit a raw sequence.

The applicability to edit a raw sequence adds another dimension to research in video retargeting. For instance, a user can simply record the scene from a static/moving camera covering the entire scene and can later use a retargeting algorithm to edit the recording based on gaze data. We motivate this application based on examples recorded from content rich theatre recordings and the user study confirms that the retargeted version better conveys the important details and significantly improves the overall viewing experience. 

The robustness of our algorithm to noise (\ie, spurious gaze data) allows us to employ gaze recordings obtained from a low cost Tobii Eyex (\euro100) eye tracker for generating the retargeted video. Such eye trackers can be easily connected to any computer and in fact may come integrated with laptops in future (https://imotions.com/blog/smi-apple-eye-tracking/), which generates the possibility of (a) creating personalized edits and (b) crowd sourcing gaze data for more efficient video editing. The current computational time of out algorithm is about 10 minutes to retarget a 90 minute video, which makes it suitable for directly editing full-length movies. 


The performed user study further confirms that our approach enables users to obtain a better view of scene emotions and actions, and in particular enhances viewing experience for static theatre recordings. Alternatively, our algorithm can also be effective for safety and security applications as it would enable a detailed view of events that capture the attention of a surveillance operator.

One limitation of our approach is that it only optimizes over the $x$-position and zoom. The $y$-position is not altered and that limits the algorithm from (a) retargeting to arbitrary aspect ratios and (b) to freely manipulate the compositions (for instance, retargeting from a long shot to a medium shot, \etc). The current version of our algorithm may result in videos where faces or body are cut by the frame boundary. However, it avoids cutting objects that are attended upon, and this  problem can be partially handled via additional constraints based on human/object detection. 

\section*{Acknowledgements} This work was supported in part by Early Career Research Award, ECR/2017/001242, from Science and Engineering Research Board (SERB), Department of Science \& Technology, Government of India and  IIIT-H seed grant award.  Special thanks to Claudia Stavisky, Auxane Dutronc and the cast and crew of `Death of a salesman' and `Cat on a hot tin roof'.

\bibliographystyle{eg-alpha-doi}

\bibliography{egbibsample}






\end{document}